%% file: iclr_main.txt
\documentclass{article} 
\usepackage{iclr2024_conference,times}

\input{math_commands.tex}

\usepackage{microtype}
\usepackage{graphicx}
\usepackage{subfigure}
\usepackage{caption}
\usepackage{booktabs} 

\usepackage{url}

\usepackage{breakurl}
\usepackage[breaklinks]{hyperref}
\definecolor{darkblue}{rgb}{0.0,0.0,0.65}

\hypersetup{
  colorlinks = true,
  citecolor  = darkblue,
  linkcolor  = darkblue,
  citecolor  = darkblue,
  filecolor  = darkblue,
  urlcolor   = darkblue,
}

\newcommand{\RI}[1]{\textcolor{blue}{\textit{#1}}}
\newcommand{\ul}[1]{\underline{#1}}
\usepackage{ulem}
\usepackage{setspace}

\usepackage{subfiles}
\usepackage{url}
\usepackage{caption}
\usepackage{graphicx}
\usepackage{amsfonts}
\usepackage{amsmath}
\usepackage{amsthm}
\usepackage{amssymb}
\usepackage{mathtools}
\usepackage{bbm}
\usepackage{utfsym}
\usepackage{algpseudocode}
\usepackage{algorithm}
\microtypesetup{nopatch={footnote}}
\renewcommand{\algorithmicrequire}{\textbf{Input:}}

\theoremstyle{definition}

\newcommand{\Sref}[1]{\S\ref{#1}}

\newcommand{\hw}{\hat{w}}

\usepackage{xcolor}

\title{LatticeGen: A Cooperative Framework which Hides Generated Text in a Lattice for Privacy-Aware Generation on Cloud}

\author{Antiquus S.~Hippocampus, Natalia Cerebro \& Amelie P. Amygdale \thanks{ Use footnote for providing further information
about author (webpage, alternative address)---\emph{not} for acknowledging
funding agencies.  Funding acknowledgements go at the end of the paper.} \\
Department of Computer Science\\
Cranberry-Lemon University\\
Pittsburgh, PA 15213, USA \\
\texttt{\{hippo,brain,jen\}@cs.cranberry-lemon.edu} \\
\And
Ji Q. Ren \& Yevgeny LeNet \\
Department of Computational Neuroscience \\
University of the Witwatersrand \\
Joburg, South Africa \\
\texttt{\{robot,net\}@wits.ac.za} \\
\AND
Coauthor \\
Affiliation \\
Address \\
\texttt{email}
}

\newcommand{\secvsabove}{\vspace{-1.1mm}}
\newcommand{\secvsbelow}{\vspace{-1.4mm}}
\newcommand{\subsecvs}{\vspace{-1mm}}
\newcommand{\figvsmid}{\vspace{-1.7mm}}
\newcommand{\figvsbottom}{\vspace{-3.4mm}}
\newcommand{\paravs}{\vspace{-2.2mm}}
\newcommand{\eqvs}{\vspace{-2.4mm}}

\begin{document}

\maketitle

\begin{abstract}
In the current user-server interaction paradigm of prompted generation with large language models (LLM) on cloud, the server fully controls the generation process, which leaves zero options for users who want to keep the generated text to themselves. We propose LatticeGen, a cooperative framework in which the server still handles most of the computation while the user controls the sampling operation. The key idea is that the true generated sequence is mixed with noise tokens by the user and hidden in a noised lattice. Considering potential attacks from a hypothetically malicious server and how the user can defend against it, we propose the repeated beam-search attack and the mixing noise scheme. In our experiments we apply LatticeGen to protect both prompt and generation. It is shown that while the noised lattice degrades generation quality, LatticeGen successfully protects the true generation to a remarkable degree under strong attacks (more than 50\% of the semantic remains hidden as measured by BERTScore). 

\end{abstract}

\secvsabove
\section{Introduction}
\label{sec:intro}
\secvsbelow



Many of the high-performing large language models (LLMs) need to be deployed on cloud servers, whether they are open-sourced but have an intensive need for computation \citep{zhao2023surveyllm,jared2020scaling}, or behind a paywall like ChatGPT \citep{openai2023gpt4}. This raises new privacy challenges \citep{li2021large,yu2021differentially,kerrigan-etal-2020-differentially}, since users have to send or receive their data to/from cloud providers.


In this work we focus on a popular interaction paradigm between end users and a server hosting an LLM on cloud named prompted generation: The user sends server a prompt, which is usually an instruction \citep{chung2022scaling} or a beginning of a document, and the server, who fully controls the generation process, sends user back the generated text from the LLM. Both the prompt and the generation are raw texts which are completely transparent and accessible to the server, leaving zero options for users who want to keep the generated text to themselves. This is a serious problem as LLMs become widely deployed in professional and social applications. \footnote{See \Sref{appsec:industry} for a more involved discussion about the current industry state.}

Existing work in privacy-aware NLP \citep{chen21bertprivacy, mcmahan17dprnn} mostly focuses on protecting user data for training (e.g., federated learning \citep{huang-etal-2020-texthide}) or inference, and the majority of works focus on natural language understanding (NLU) tasks \citep{feyisetan20privacy}. We argue that in prompted generation, there are many scenarios in which user prompts \textit{as well as the generated contents} need obfuscation, because they can directly affect the user's decisions (as we discuss in \Sref{sec:preliminary}).

With the goal of preventing the server from gaining complete knowledge of the generated text and prompt, we propose LatticeGen (Figure \ref{fig:latticegen}), a user--server interaction framework in which the user and server conduct prompted generation token-by-token in \textbf{a cooperative way}. We summarize our key contributions below:

\begin{itemize}
    \item The key idea of LatticeGen (\Sref{sec:approach}) is that in each time-step, the user sends the sever not one, but $N$ tokens (thus the name \textit{lattice}), in which one is true and others act as noise. The server does LLM inference and sends user back the next-token prediction distributions for all $N$ tokens, which are used by user to sample the true and noise tokens for the next time-step.
    
    \item Considering potential attacks from a hypothetically malicious server and how the user can defend against it (\Sref{sec:atkdef}), we propose the repeated beam-search attack and the mixing noise scheme.

    \item We apply LatticeGen to the task of creative writing \citep{fan2017controllable-topk}. Our experiments (\Sref{sec:exp}) show that while the noised lattice degrades generation quality, LatticeGen successfully prevents a malicious server from recovering the true generation to a remarkable degree (more than 50\% of the semantic remains unknown as measured by BERTScore). \footnote{Our code and data will be available in the public version of this manuscript.}
\end{itemize}

\input{sec_lattice}

\input{sec_atkdef}

\input{sec_exp}

\input{sec_discuss}

\input{sec_related}

\secvsabove
\section{Conclusion}
\secvsbelow

LatticeGen aims for an ambitious and seemingly conflicting goal: The server still does most computation for the generation but does not know what exactly is generated. This is achieved by our proposed noised lattice structure, and a cooperative generation protocol between the server and user. 

While the noised lattice degrades generation quality and inference speed, LatticeGen with our proposed mixing noise scheme successfully prevents a malicious server from recovering the true generation to a remarkable degree (more than 50\% of the semantic remains unknown as measured by BERTScore). We hope our work could inspire more research into this under-studied yet important field of privacy-aware LLM generation on cloud.



\normalem \bibliography{custom,anthology}
\bibliographystyle{iclr2024_conference}

\newpage
\input{appendix.tex}

\end{document}

%% file: math_commands.tex
\usepackage{amsmath,amsfonts,bm}

\def\eqref#1{equation~\ref{#1}}

\def\1{\bm{1}}

\DeclareMathAlphabet{\mathsfit}{\encodingdefault}{\sfdefault}{m}{sl}
\SetMathAlphabet{\mathsfit}{bold}{\encodingdefault}{\sfdefault}{bx}{n}

\DeclareMathOperator*{\argmax}{arg\,max}

%% file: sec_lattice.tex
\secvsabove
\section{Motivation and Preliminaries}
\label{sec:preliminary}
\secvsbelow


\subsection{Generated Text (also) Needs Obfuscation}
\subsecvs

In the current user--server interaction paradigm, the user sends the server a prompt which is usually the beginning of a dialogue, story or instruction, then the server generates a complete response autoregressively (\Sref{sec:preliminary}), and sends it back to the user. Both the prompt and generation are directly available to the server in raw text format.

This paper contends that generated texts, as well as user prompts, require a privacy protection mechanism.  
A key reason is that in various scenarios, the generation from the LLM can affect the user's private decisions: e.g., a customer is likely to go to the restaurant suggested by the LLM; a writer could take inspiration from outputs provided by the LLM; an engineer or manager could adopt the approach proposed by the LLM. Industry regulations do not provide ample protection. Please see \Sref{appsec:industry} for recent privacy-related incidents with ChatGPT or Bard. 
The goal of our LatticeGen protocol is to provide a controlled level of obfuscation for the generated text, making it difficult for a hypothetically malicious server to infer the user's actions after interacting with the LLM. 



\subsecvs
\subsection{LatticeGen as a Third-Party Client}
\subsecvs

\begin{figure}[t]
    \begin{center}
    \vspace{-5mm}
    \includegraphics[width=\linewidth]{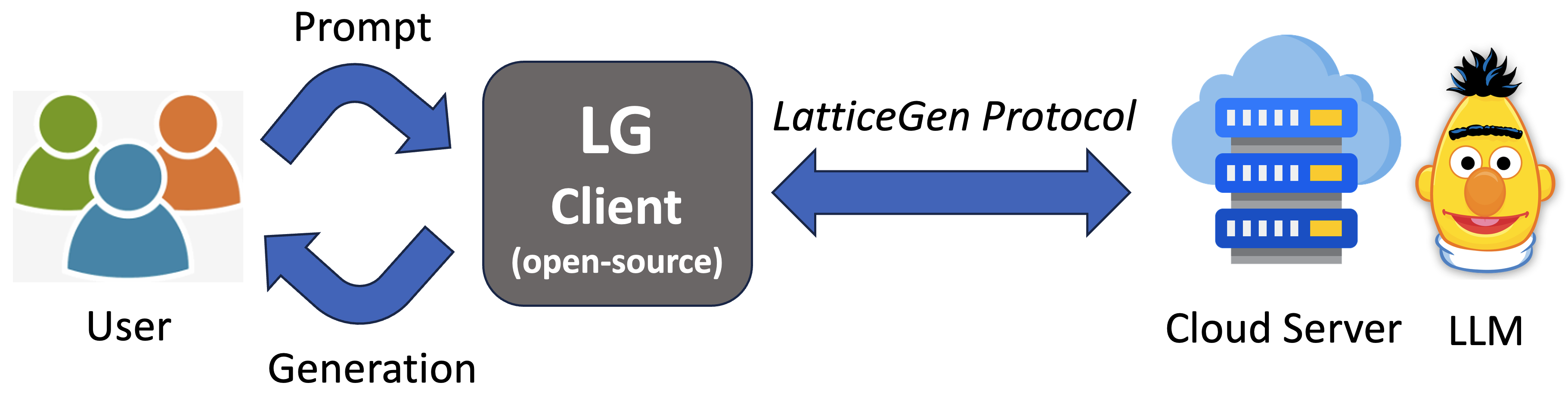}   
    \end{center}
  \vspace{-4mm}
  \caption{LatticeGen can be implemented as a third-party client handling the protocol for the user.} 
  \vspace{-4mm}
  \label{fig:client}
\end{figure}

Before expanding on the proposed protocol (\Sref{sec:approach}), we first clarify that LatticeGen does not complicate the user interface. Indeed, it is likely that most users still want to keep a simple and intuitive interface for prompted generation. In light of this, LatticeGen can be implemented as a third-party client between the user and the server. As Figure \ref{fig:client} depicts, the client takes the prompt from the user, conducts the privacy-aware generation protocol with the server, and finally returns the generation to the user. In this way, the user does not need to deal with the complicacy in the protocols. 

The next question is why would a common user trust the client? One solution is that the client can be open-sourced (e.g., as python scripts) and therefore vetted by researchers and users worldwide. It can also facilitate comprehensive evaluations conducted by different research groups. The user only need to download the script and set the hyper-parameters (e.g., random seed).

\begin{figure*}[t]
    \centering
    \includegraphics[width=\linewidth]{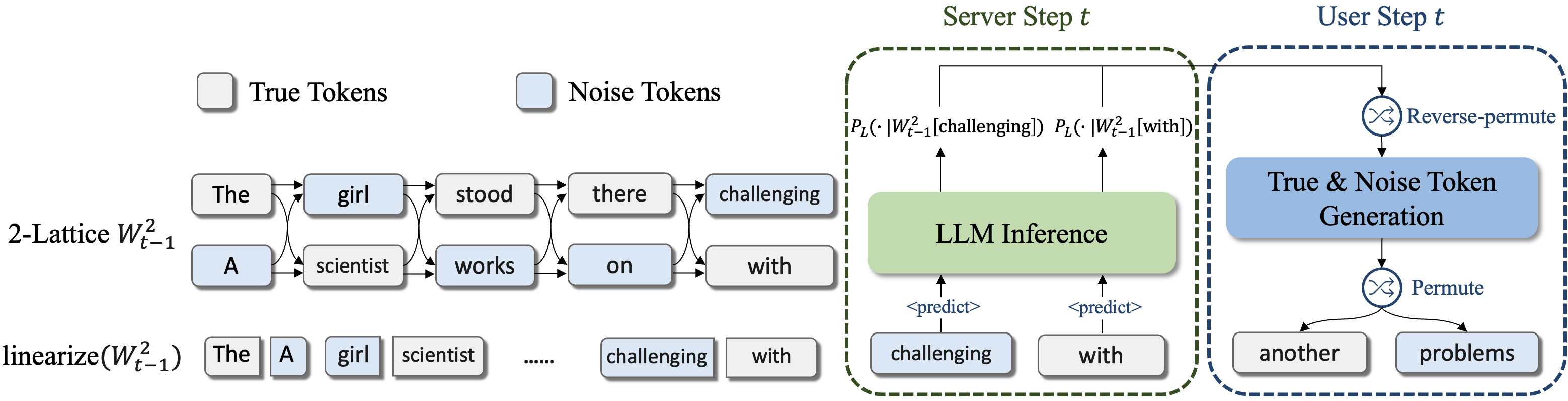}
    \figvsmid
    \caption{Client-Server interaction under LatticeGen for time-step $t$. The server controls the LLM $P_L$, conducts the inference computation and sends client the next-token prediction distribution for each received token. The client conducts the sampling of the true and noise token(s), and sends server a randomly permutated list of tokens for the next time-step. \textbf{The server does not know which tokens are the true ones.} The task is creative writing, and the prompt part is omitted in this figure for brevity. An illustration of the server step for $N=3$ and $G=2$ is provided in Figure \ref{fig:serverstep_bigram}, Appendix \ref{appsec:framework}.}
    \label{fig:latticegen}
    \figvsbottom
\end{figure*}

\subsecvs
\subsection{Preliminaries}
\label{sec:preliminary}
\subsecvs

We will start by reviewing the traditional autoregressive LM generation, and then move on to introduce necessary components of LatticeGen.

\paravs
\paragraph{Traditional Autoregressive LM Generation} We assume the server-side LLM is an autoregressive LM, i.e., it generates tokens one at a time and from left to right \citep{mikolov2012statistical,cho-etal-2014-learning,huszár2015not,Welleck2020Neural,transformerxl,nitish2019ctrl}. We denote the LLM as $P_M$ with parameter set $\theta$, the vocabulary as $V$, the generated token at time-step $t$ as $w_t$, and the given prompt as $p$. For convenience we regard the prompt as part of generation, therefore, $w_t:=p_t$ for $1 \leq t \leq \text{len}(p)$. In traditional autoregressive generation, on each time-step $t > \text{len}(p)$, the next token $w_{t}$ is sampled from $P_M(\cdot | w_{0..t-1})$ by calling a sampling algorithm such as top-$k$ \citep{fan2017controllable-topk} or nucleus sampling \citep{Holtzman2020toppnucleus}. $w_0$ is the \texttt{<bos>} token.

\paravs
\paragraph{The Lattice Structure} A simple but key concept in our proposed framework is the \textit{lattice}. In a width-$N$ lattice (or an $N$-lattice for short), each time-step contains $N$ token options and we denote them as $\{w^1_t, ... , w^N_t\}$. Therefore, a $N$-lattice of length $T$ (denoted as $W^N_T$) represents $N^T$ possible sequence combinations. An example with $N=2$ is shown in the left part of Figure \ref{fig:latticegen}. 


In our proposed LatticeGen protocols (\Sref{sec:protocols}), for each time-step $t$, only the client knows which token is the ``true'' one, denoted by $\wtt_t$. And the other $N-1$ tokens $\{w^\text{noise(1)}_t, ... , w^\text{noise(N-1)}_t\}$ are referred to as ``noise'' tokens. Therefore we will also refer to it as the \textit{noised lattice}. To prevent the server from knowing which one is the true token, the client will randomly shuffle the list before attaching it to the lattice and sending to the server. 

\paravs
\paragraph{LM Finetuning and Inference with the LLG (Linearized Lattice plus G-gram) Format} As a prerequisite for LatticeGen, we need the server-side LLM \citep{tfattention17Vaswani} to be able to do inference based on a given lattice and we achieve that by finetuning the base LLM $P_M$ to make next-token prediction with the LLG (Linearized Lattice plus $G$-gram) format. Below we first introduce this format, and describe the finetuning objective.

First, as the name suggests, we conduct a simple \textit{linearization} operation before feeding the lattice to the LM, in which the token options on each time-step are linearized and concatenated into a sequence of length $T\times N$ (see Figure \ref{fig:latticegen} for an example): 
\begin{equation}
\small
\text{linearize}(W^N_T)=[\texttt{<bos>}] + \text{concat}^T_{i=1}([w^1_i,...,w^N_i]).
\end{equation}
An illustration of a linearized lattice is given in Figure \ref{fig:latticegen}.

Next, we append a \texttt{<predict>} token and $G$ tokens specifying the token options for the last $G$ tokens (for time-step from $T-G$ to $T-1$), and the LLM is trained to predict the next token with this specified $G$-gram ``tail''. We use notation $S$ to denote a $G$-gram, where $S_i \in \{w^1_{T-G+i},...,w^N_{T-G+i}\}$ for $1\leq i \leq G$. In Figure \ref{fig:latticegen}, we use uni-gram ($G=1$) and the last token could be ``challenging'' or ``with''. The generation quality will be better with larger $G$ (since the token history is less noised), at the price of more computation: The server will need to enumerate $N^G$ potential combinations.

In \Sref{sec:modeltrain}, we describe a simple process to finetune a LLM to predict the next token for the LLG format. Here we provide a high-level description. For each data sample $w^\text{data}$, we construct and linearize a noised lattice by using $N-1$ other random data samples as noise. The LLM is then finetuned to predict the next true token for several randomly picked tokens in the data sample with the LLG format. We denote the LLG-finetuned LLM as $P_L$, and the prediction distribution for $w_t$ with a noised lattice $W^N_{t-1}$ and a specific $G$-gram tail $S$ as $P_L(\cdot|W^N_{t-1}[S])$. In most parts of this paper, we will assume unigram ($G=1$) just for notation simplicity.



\secvsabove
\section{LatticeGen}
\label{sec:approach}
\secvsbelow

To prevent the server from gaining full knowledge of the generation and prompt, LatticeGen makes several core changes to the client--server interaction. On a high level, the server who possesses the LLG-finetuned LLM $P_L$ (the finetuning is detailed in \Sref{sec:modeltrain}) still handles most of the computation, while the client controls the token sampling operations and expands the lattice to the next time-step. In particular, the client will sample one true token and $N-1$ noise tokens, where $N \geq 2$ is a hyperparameter controlling the width of the lattice. \textbf{In the end, both the server and client obtain the same noised lattice $W^N_{T}$, but only the client knows which token is the true one for each time step.}

In the beginning, the server needs to share the vocabulary $V$ with the client, but all other parameters or configurations of the LLM are not shared. We describe the protocol below. 






\subsecvs
\subsection{Protocol}
\label{sec:protocols}
\subsecvs

For simplicity, we first ignore the prompt part and assume the generation starts at the first token. In the beginning $t=0$, both the server and client begin with an empty local lattice, and the client sends $N$ $\texttt{<bos>}$ tokens to the server. We divide the client--server interaction at each time-step $t \geq 1$ into a \textit{server step} and a \textit{client step}, illustrated by Figure \ref{fig:latticegen} (also see Algorithm \ref{alg:latticegen}). 


\paravs
\paragraph{Server Step} From the last time-step, the server receives from client $N$ tokens $\{w^1_{t-1},...,w^N_{t-1}\}$ and expands its local lattice to $W^N_{t-1}$. The server does not know which received token is the true token because the list is shuffled by the client, and computes the respective next-token prediction distribution for all $N^G$ potential $G$-gram tails with the LLG format (each potential tail is denoted as $S$). More concretely, the lattice $W^N_{t-1}$ is linearized, appended with each $G$-gram, and fed to $P_L$, which outputs $\{P_L(\cdot|W^N_{t-1}[S^i])\}^{N^G}_{i=1}$. \footnote{In the uni-gram case, the notation simplifies to $\{P_L(\cdot|W^N_{t-1}[w^i_{t-1}])\}^{N}_{i=1}$.}

Since all $G$-grams share the same linearized lattice, the inference can be made efficient by reusing transformer hidden states and parallel computing. We defer the details of finetuning and inference (both conducted by the server) to \Sref{sec:modeltrain}.  The server represents the next-token prediction distributions as $N^G$ length-$|V|$ vectors, and sends them back to the client.

\paravs
\paragraph{Client Step} Different from the server, the client knows which tokens are the true ones. Upon receiving the list of distribution vectors from the server, the client  extracts the distribution for the true $G$-gram $P_L(\cdot|W^N_{t-1}[\wtt_{(t-G)...(t-1)}])$, from which the client samples $\wtt_t$. The client also need to generate $N-1$ ``noise'' tokens $\{w^\text{noise(1)}_{t}, ... , w^\text{noise(N-1)}_{t}\}$ with a certain noise scheme. 

How to generate noise tokens is a key part of making the noised lattice robust to potential attacks from the server side. For now, we assume a simple synonym noise scheme in which we use synonyms of the true token. Concretely, $w^\text{noise}_t$ is randomly sampled from $S$ tokens having the closest embedding with $\wtt_t$ measured by cosine similarity. In our experiments we set $S=5$. \footnote{In practice, we exclude the first ten closest token in $V$, as their surface forms are usually very close to the true token, making the obfuscation useless (e.g., only different in capitalization).}
In practice this simple noise scheme will be vulnerable to attacks from a malicious server. See \Sref{sec:atkdef} for discussions on attacks and our proposed advanced noise schemes for defense. 

 
With a private random seed, the client randomly permutates the token list and sends it to the server. This concludes the client--server interaction in time-step $t$. 

\paravs
\paragraph{Incorporating Prompts (Client)} The incorporation of prompts is quite straightforward by regarding it as a prefix of the generation, and the content in the prompt can also be noised and protected by LatticeGen. See \Sref{appsec:prompt} for implementation details.


\renewcommand{\algorithmicrequire}{\textbf{Input}}
\renewcommand{\algorithmicensure}{\textbf{Output}}

\begin{algorithm}[t] \scriptsize 
\caption{\footnotesize Pseudo-code for LatticeGen}\label{alg:latticegen}
    \begin{algorithmic}
        \Require \textbf{(Server):} Lattice-finetuned LLM $P_L$, lattice width $N$, generation length $T$, and inference tail length $G$.
        \Require \textbf{(Client):} Prompt $p$, a noise generation scheme, a private large prime number for random seed.

        \State Client sets $w^i_0:=\texttt{<bos>}$ for $1\leq i \leq N$. 
        \State Both the server and client begin with an empty lattice.
        \State The client sends $[w^1_0,...,w^N_0]$ to server indicating the beginning of generation.
        
        \For{$t = 1 \dots T$}
        \State \texttt{\textcolor{blue}{\# Server Steps Below}}
        \State Receives $[w^1_{t-1},...,w^N_{t-1}]$ from client and use it to extend the lattice to $W^N_{t-1}$.
        \State For each $G$-gram tail $S^i$, run next-token inference on $P_L$ with the LLG format and obtain $\{P_L(\cdot|W^N_{t-1}[S^i])\}^{N^G}_{i=1}$.
        \State Send the distributions to the client as $N^G$ length-$|V|$ vectors.

        \State \texttt{\textcolor{blue}{\# Client Steps Below}}
        \State Receives the next-token distributions $\{P_L(\cdot|W^N_{t-1}[S^i])\}^{N^G}_{i=1}$ from server.
        \If{$t \leq \text{len}(p)$}
        \State Set $\wtt_{t} := p_{t}$.
        \Else
        \State Sample $\wtt_{t}$ from $P_L(\cdot|W^N_{t-1}[\wtt_{(t-G)...(t-1)}])$.
        \EndIf
        \State Generate $N-1$ noise tokens $\{w^\text{noise(1)}_{t}, ... , w^\text{noise(N-1)}_{t}\}$ with the noise scheme.
        \State Set the current private random seed to be $t$ multiplied by the private prime number.
        \State Obtain the permuted list $[w^1_t,...,w^N_t]$ using the current random seed.
        \State Extend the local lattice, and send $[w^1_t,...,w^N_t]$ to the server.
        \EndFor 

        \Ensure \textbf{(Server):} Lattice $W_T^N$.
        \Ensure \textbf{(Client):} True sequence $\{\wtt_t\}^N_{t=1}$, and lattice $W_T^N$.

    \end{algorithmic}
\end{algorithm} 

 We summarize the LatticeGen protocols as pseudo-code in Algorithm \ref{alg:latticegen}. The discussion on the network communication cost between client and server is deferred to \Sref{appsec:communicate} to save space.

\subsecvs
\subsection{Comparison with Standard LM: History Noised While Locally Sharp}
\label{sec:comparestandard}
\subsecvs

It is helpful to formulate a comparison between LatticeGen ($P_L$) and generation from a standard autoregressive LM $P_M$. For simplicity, we ignore the noise generation (i.e., lattice-building) part, and only care about how the true tokens are generated with $P_L$. Under this simplification, the probability of generating a true sequence $w$ is:



\eqvs
\begin{equation}
\small
\label{eq:formulate_noise}
    \log P_L(w) \approx \sum^T_{t=1} \log P_L(w_t|W^N_{t-1}[w_{(t-G)...(t-1)}]),
\end{equation}
\eqvs

where the forming process of $W^N_{t-1}$ (noise tokens and permutation) at each time-step is omitted.

For comparison, the log-probability of generating $w$ with the standard model $P_M$ is:

\eqvs
\begin{equation}
\small
\label{eq:formulate_standard}
    \log P_M(w) = \sum^T_{t=1} \log P_M(w_t|w_{0...t-2}, w_{t-1}).
\end{equation}
\eqvs

Comparing the above two equations with similar structure, it should be clear that what LatticeGen does is essentially blurring the token history $w_{0...t-2}$ by the noised lattice $W^N_{t-2}$. Therefore, increasing the number of noise tokens gives better protection for the true token sequence, but at the same time degrades the LM's performance.

\textbf{While the history is blurred, the local sharpness \citep{khandelwal-etal-2018-sharp} is preserved by LatticeGen:} From Equation \ref{eq:formulate_noise}, the exact last $G$ tokens is provided to the model. Therefore, in the worst-case scenario (zero utilization of non-immediate history), LatticeGen is at least as strong as a $(G+1)$-gram LM. 

%% file: sec_atkdef.tex
\secvsabove
\section{Attack and Defense}
\label{sec:atkdef}
\secvsbelow

In this section, we discuss potential attack algorithms from a hypothetically malicious server to decode the true token sequence $\{\wtt_t\}^T_{t=1}$ hidden in the lattice $W^N_T$, and the client's noise generation schemes as defense. For notational simplicity, we will assume unigram ($G=1$), and the extension to $G>1$ should be straightforward. We first establish metrics to measure the strength of attacks.


\paravs
\paragraph{Metrics} Given a lattice $W^N_T$, the attacker's target is to decode a hypothesis sequence $\hw$ with $\hw_t \in \{w^1_t,...,w^N_t\}$ having biggest overlap with the true generation $\wtt$. We define a simple \textit{true-ratio} metric to measure the strength of the attack: 

\eqvs
\begin{equation} \small
    \text{true-ratio}(\hw, \wtt) = \frac{\sum^T_{t=1}\mathbbm{1}_{\hw_t = \wtt_t}}{T}.
\end{equation}
\eqvs

In the repeated beam search attack to be described below, the result of the attack algorithm is not only one but $N$ sequences $\{\hw^i\}^N_{i=1}$ which spans the whole lattice (i.e., $\{\hw^i_t\}^N_{i=1}=\{w^i_t\}^N_{i=1}$). In this case, we argue that the defending noise scheme should prevent \textit{any} of the hypothesis from having a high overlap with the true sequence, and measure it with the \textit{max-true-ratio}: \footnote{The average of the true-ratio will always be $\frac{1}{N}$ because each true token is in one of the $N$ hypotheses.}

\eqvs
\begin{equation} \small
    \text{max-true-ratio}(\{\hw\}^N_{i=1}, \wtt) = \max_i \frac{\sum^T_{t=1}\mathbbm{1}_{\hw^i_t = \wtt_t}}{T}.
\end{equation}
\eqvs

\textbf{It should be clear that $\frac{1}{N}$ is a lower bound for max-true-ratio for any noise scheme, which provides an intuition of why larger $N$ would better protect the true sequence.}

Albeit intuitive, a big weakness of the true-ratio metric is that it only considers exact matches and does not reflect the semantic similarity between the hypothesis and the true generation. Therefore, in our experiments we will also use an embedding-based metric BERTScore \citep{Zhang2020BERTScore} to measure the leaked information on semantics. Similar to true-ratio, BERTScore is larger than zero and has a maximum value of 1 (we refer readers to its paper for details). We define max-BERTScore in the same fashion as max-true-ratio and we omit the formulation for brevity.


\begin{figure}[t]
    \centering
    \includegraphics[width=\linewidth]{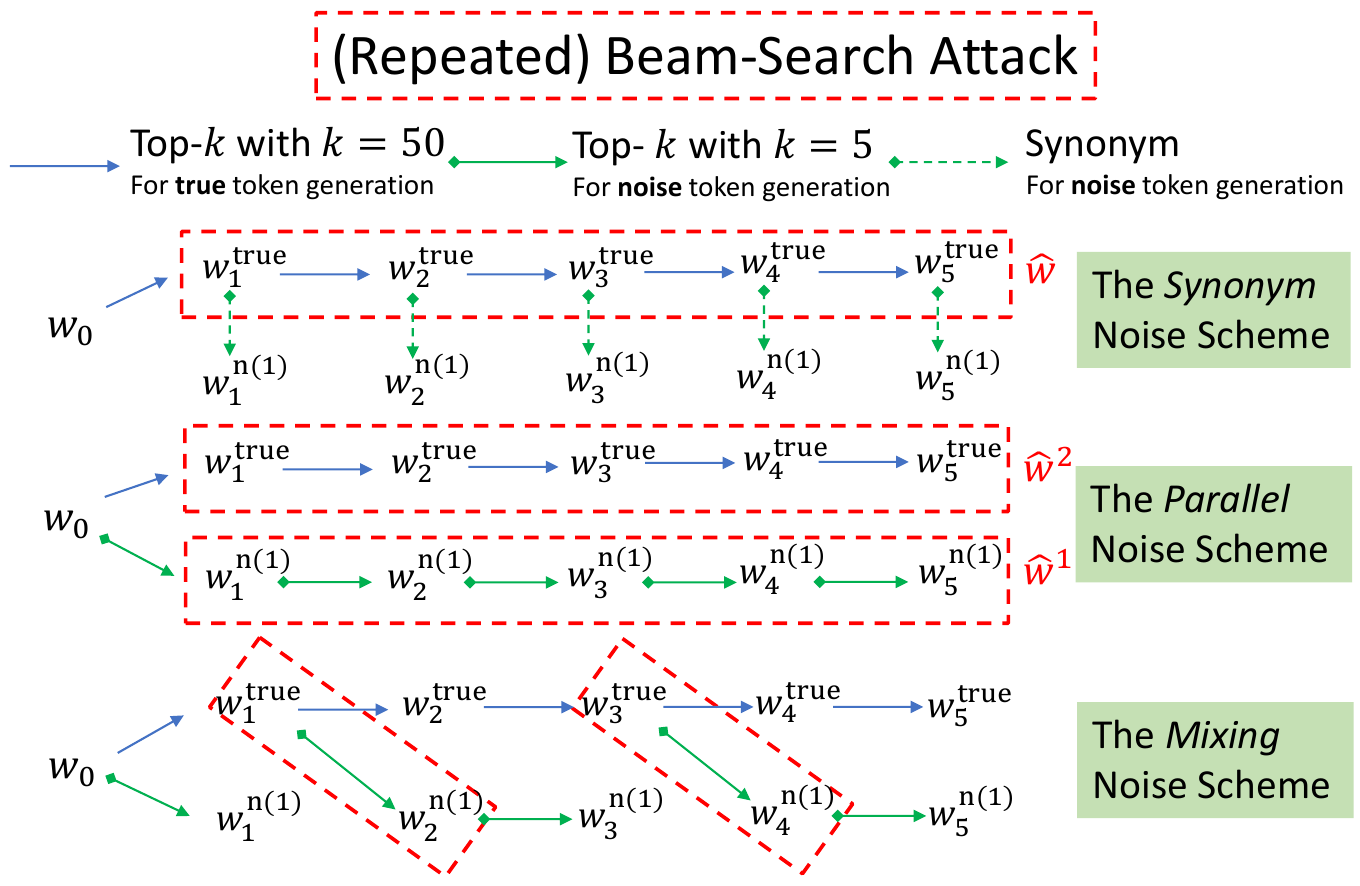}
    \figvsmid
    \caption{Illustration of different noise schemes under (repeated) beam-search attack. For convenience, the lattice is not shuffled on each time-step. An illustration with a width-3 lattice is given in Figure \ref{fig:rbsatk_n3} (\Sref{appsec:framework}).}
    \label{fig:rbsatk}
    \figvsbottom
\end{figure}

\subsecvs
\subsection{The Repeated Beam-Search Attack}
\label{sec:atk_beamsearch}
\subsecvs


In this section, we motivate and describe the \textit{repeated beam-search attack} which is the major attack algorithm considered in this work. It is a stronger version of the \textit{beam-search attack} described below.

\paravs
\paragraph{The Beam-Search Attack (Server)} Assuming unigram unit, a natural objective for the attacker is to find the sequence $\hw$ with $\hw_t \in \{w^1_t,...,w^N_t\}$ which is mostly likely to be generated by $P_L$:

\eqvs
\begin{equation} \footnotesize
\begin{aligned}
    \argmax_{\hw} & \log P_L(\hw|W^N_T) = \\
    & \argmax_{\hw} \sum^T_{t=1} \log P_L(\hw_t|W^N_{t-1}[\hw_{t-1}]).
\end{aligned}
\end{equation}
\eqvs

By saving all probability distributions during the generation, the attacker can efficiently conduct this optimization using the classical beam-search algorithm. We term it as the \textit{beam-search attack}.


Our experiments (\Sref{sec:exp}) show that the simple synonym noise scheme discussed in \Sref{sec:approach} is highly vulnerable to the beam-search attack. We show some intuition in the upper part of Figure \ref{fig:rbsatk}: There does not exist a direct link between the noise tokens. The log-probability of the true sequence will likely be much higher than any combination of the noise tokens, and is therefore revealed by the attack.

\paravs
\paragraph{The Parallel Noise Scheme (Client)} There is an intuitive way to defend against the beam-search attack: The client can sample a noise sequence independent of the true sequence, and make it have higher log-probability than the true sequence by tuning the hyper-parameter of the sampling algorithm. We term it the \textit{parallel noise scheme} and illustrate in the middle of Figure \ref{fig:rbsatk}.

More concretely, at time-step $t$, the $i$-th noise token is sampled from $P_L(\cdot|W^N_{t-1}[w^\text{noise(i)}_{t-1}])$. \footnote{If $G>1$, the last $G$ tokens from the $i$-th the noise sequence will be used.} In this way, the noise sequences $w^\text{noise(i)}$ are parallel and independent of the true sequence $\wtt$. We also assume the adoption of popular sampling hyper-parameter for the generation of the true sequence (e.g., $k=50$ for top-$k$ or $p=0.96$ for nucleus), which enables the adoption of a more radical hyper-parameter \cite{Caccia2020Language,nadeem-etal-2020-systematic} for the sampling of the noise sequences: in our experiments we use $k=5$.    

Our experiments show that the parallel noise sequences can very effectively hide the true sequence from the beam-search attack. This motivates our proposed repeated beam-search attack.     

\paravs
\paragraph{The Repeated Beam-Search (RBS) Attack (Server)} We propose a simple but more powerful attack algorithm based on the beam-search attack: Given a $N$-lattice, we do beam-search $N-1$ times. After obtaining the resulting hypothesis sequence of the $i$-th  beam-search (denoted as $\hw^i$), we remove the tokens in $\hw^i$ from the lattice, resulting in a $(N-i)$-lattice. After the $(N-1)$-th beam-search, only one sequence is left in the lattice, which becomes the $N$-th hypothesis $\hw^N$. We term it the repeated beam-search (RBS) attack. 

The intuition of why the RBS attack is effective against the parallel noise scheme is shown in the middle of Figure \ref{fig:rbsatk}. Since the noise sequences are of high probability and independent of each other, it is likely that the $N-1$ times of beam-search would obtain all the noise sequences as hypotheses which are removed from the lattice in turn, and the remaining true sequence is therefore revealed in the end as $\hw^N$. This would result in a high max-true-ratio. 

\subsecvs
\subsection{The Mixing Noise Scheme for Defense}
\label{sec:def_branch}
\subsecvs

We propose the \textit{mixing noise scheme} to defend against the RBS attack, with the intuition that the true and noise sequences should somehow be mixed. This scheme can be regarded as a variant of the parallel noise scheme. Again we adopt a radical hyper-parameter for the sampling of the noise sequences (top-$k$ with $k=5$). At time-step $t$, with a random ratio determined by a hyper-parameter \textit{mix-ratio}, the $i$-th noise token is sampled from $P_L(\cdot|W^N_{t-1}[\wtt_{t-1}])$, \textbf{which is the next-token distribution for the true sequence}. \footnote{We will re-sample if the sampled token is the same as the true token.} Otherwise we sample from $P_L(\cdot|W^N_{t-1}[w^\text{noise(i)}_{t-1}])$, same as in the parallel scheme. 

We illustrate this at the bottom of Figure \ref{fig:rbsatk}. In comparison to the parallel scheme, the goal is to make the sequence with the highest log-probability be a mix between the true and noise sequences. And the key is to make the true sequence ``branch'' out to the noise sequences, which breaks the continuity of the noise sequences. Although broken, the radical sampling used for the noise sequence would still attract the repeated beam-search attack, and the true and noise sequences are mixed by the branching connections. Our experiments show that with a tuned mix-ratio, the mixing noise scheme achieves the best max-true-ratio under RBS attack.

%% file: sec_exp.tex
\secvsabove
\section{Experiments}
\label{sec:exp}
\secvsbelow

\input{main_table}

\subsection{Experiment Setting}
\label{sec:exp_setting}

\paragraph{Model \& Noise Schemes} We use the OPT-1.3B \citep{zhang2022opt} and the Llama2-7B model as our base LLM, from which both $P_L$ and $P_M$ are finetuned. We select those models due to limited computing resource and as a proof-of-concept. Our protocol can be readily applied to larger autoregressive LMs such as GPT3 or GPT4. In our implementation, for convenience we simulate the client--server interaction protocols on a single machine. 

For sampling of the true sequence, we use top-$k$ \citep{fan2017controllable-topk} sampling with $k=50$, temperature 0.7, and a repetition penalty of 1.05. For the noise token sampling in the parallel or mixing noise scheme, $k=5$ is used. It should be clear that LatticeGen can also be applied to other sampling algorithms with proper hyper-parameters. We limit the maximum generation length to 60 tokens. For the mixing noise scheme of OPT, we use a mix-ratio of 0.1 for both $N=2$ and $N=3$ for the generation part. For the prompt part, we use a mix-ratio of 0.2.
For Llama2, we use a mix-ratio of 0.05 for both $N=2$ and $N=3$ for the generation part and 0.2 for the prompt part. They are found to achieve the lowest max-true-ratio on the dev set.

\paravs
\paragraph{Dataset \& Lattice Finetuning} Since the word history is noised (discussed in \Sref{sec:comparestandard}), LatticeGen is not recommended for tasks with high requirements for consistency or factuality \citep{pagnoni-etal-2021-understanding}. In this work we focus on the task of creative writing \citep{martin17storygen,lili18planwrite,fan-etal-2019-strategies}, and utilize the WritingPrompts dataset \citep{fan-etal-2018-hierarchical}. The dataset is composed of stories and the corresponding high-level descriptions as prompts. The average length of prompts/stories is 29/674. We use 200/500 samples from the valid/test set for development/evaluation. The training set (10,000 samples) is used for  finetuning of $P_L$ and $P_M$, and we defer details to \Sref{sec:modeltrain}.

\paravs
\paragraph{Metrics} We use a larger LLM, namely OPT-2.7B or Llama2-13B, to measure the generation's quality or alignment with the prompt. For quality, we use the popular perplexity metric. For alignment, we use pointwise mutual information (PMI) \citep{shi2023trusting}:
\begin{equation} \footnotesize
    \text{PMI}_\text{OPT}(x;y) = \frac{\log P_\text{OPT}(x|y) - \log P_\text{OPT}(x)}{\text{len}(x)},
\end{equation}
where $x$ and $y$ denote the generation and prompt. 

To compare between different noise schemes and measure the (semantic) overlap between the attack hypothesis ($\hw$) and the true sequence ($\wtt$) under RBS attack, we use the true-ratio or BERTScore discussed in \Sref{sec:atkdef}. We will report true-ratio for the BS attack and max-true-ratio under RBS attack, and the same applies to BERTScore.

\input{figure_main}

\subsecvs
\subsection{Experiment Results}
\subsecvs

\begin{figure}
    \centering
    \includegraphics[width=0.9\linewidth]{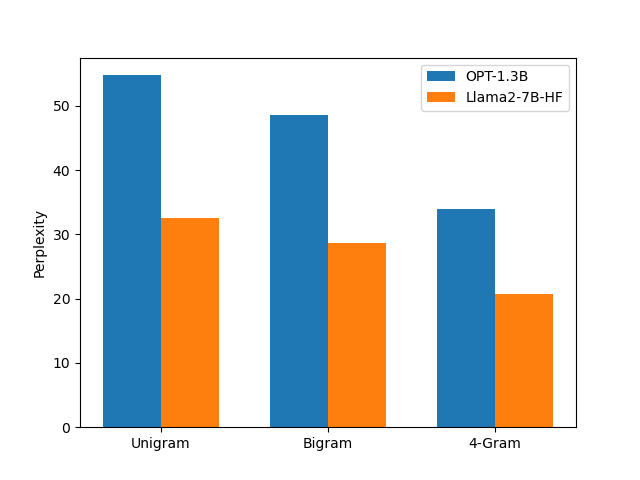}
    \figvsmid
    \figvsmid
    \caption{Comparison of perplexity of OPT-1.3B and Llama-7B-HF models on various G-gram units.}
    \label{fig:ppl-compare}
    \figvsbottom
\end{figure}

Table \ref{tab:main_gen} includes the main results when LatticeGen (LG) is applied to both generation and prompt. The standard vanilla model ($P_M$) enjoys the best generation quality (PPL and PMI), while having zero obfuscation (100\% true-ratio).

LatticeGen sacrifices generation quality (due to noised history) for obfuscation. The empirical behavior of the three noise schemes aligns with their respective intuitions discussed in \Sref{sec:atkdef}: The synonym scheme is completely defenseless against the BS attack; The parallel scheme is most effective against BS with true-ratio lower than 20\%, but is vulnerable under the stronger RBS attack.

The mixing scheme, which is our main recommended scheme, achieves the best protection under the RBS attack. For $N=2$, The max-true-ratio/BERTScore is close to or lower than 65\%/55\%. \textbf{It indicates that around half of the semantic is hidden from the attacker, and is close to the theoretical best max-true-ratio ($\frac{1}{N}=50\%$).} The protection is better with $N=3$ (50\%/40\%), but with worse generation quality.

Comparing to unigram unit, \textbf{the quality degradation (especially PPL) is alleviated to a large degree by using 4-gram units} (See Figure \ref{fig:ppl-compare} for a comparison). One could also try larger $G$-gram for further improvement. However, the computational cost would grow exponentially and we leave it to future work due to limited resources. 

What if we directly apply noise to generation but \textit{without the lattice structure}? We add an additional non-lattice baseline with the same synonym scheme used in LatticeGen: On every time-step, the client gets next-token distribution from the server and generates a true token, but sends a synonym of it back to the server. The finetuning is modified accordingly with details given in \Sref{appsec:nonlattice_baseline}. 

As shown in Table \ref{tab:main_gen}, we apply the synonym scheme to 100\% or 50\% of the tokens. The synonym noise without lattice results in drastically degraded PPL and PMI. In comparison, LatticeGen provides a trade-off between quality degradation and privacy protection. This implies that \textbf{for decent generation performance, the true tokens have to be revealed to the server in some way}. 


Table \ref{tab:speed_table} (\Sref{appsec:auxres}) compares generation speed of different systems. On the single A40 GPU we use, LG with 4-gram ($N=2$) units has a 4.76 times slowdown comparing to $P_M$. Since inference with transformer model benefits from parallel computing, the slowdown should be less significant on servers with stronger computing power. 

We show a generation example with RBS attack outputs in Figure \ref{fig:mainexample}. LG is able to generate a sample with decent quality. More importantly, around half of the story semantics remains hidden from the RBS attack by the mixing noise scheme. More examples and analysis are deferred to \Sref{appsec:auxres}.


%% file: main_table.tex
\begin{table*}[t]
\footnotesize
\centering
\addtolength{\tabcolsep}{-4.0pt}
\resizebox{0.95\textwidth}{!}{
\begin{tabular}{l|cccccc|cccccc}
\toprule
Config  & & \multicolumn{4}{c}{$N=2$ (LG only)}  & & \multicolumn{6}{c}{$N=3$ (LG only)} \\
\midrule
\multicolumn{1}{r|}{Metric} &    PPL      &    PMI    & \multicolumn{2}{c}{True-Ratio} & \multicolumn{2}{c|}{BERTScore} &    PPL      &    PMI    & \multicolumn{2}{c}{True-Ratio} & \multicolumn{2}{c}{BERTScore}   \\
   \multicolumn{1}{r|}{Attack}       &       &   & BS            & RBS           & BS           & RBS           &      &  &  BS          & RBS           & BS   & RBS   \\ \midrule
OPT, Vanilla ($P_M$), w.o. noise      & 21.272 & .345  & 1.0           & 1.0         & 1.0 & 1.0  & /          & /    & / & /  & / & /   \\ 
OPT, Synonym, w.o. lattice     & 229.616 & .058  & /          & /         & / & /  & /          & /     & / & / & /  & /  \\ 
OPT, Syn-50\%, w.o. lattice     & 199.621 & .058  & /          & /         & / & /  & /          & /     & / & / & /  & /  \\ 
OPT, LG, 4-gram, synonym & 37.574    & .244   & .993            & .993 & .894          &  .894    & 41.379   & .221 & .985            & .985 & .882&.882 \\
OPT, LG, 4-gram, parallel & 33.907     & .228   & .168            & .844 & .234          &  .784    & 35.691   & .232 & .110            & .749 & .155 & .676 \\
OPT, LG, 4-gram, \textbf{mixing} & 34.058     & .219   & .541            & \textbf{.651} & .432          &  \textbf{.531}    & 35.910   & .242 & .357            & \textbf{.511} & .285 & \textbf{.393} \\  \midrule
Llama2, Vanilla ($P_M$), w.o. noise      & 14.710 & .785  & 1.0           & 1.0         & 1.0 & 1.0  & /          & /    & / & /  & / & /   \\ 
Llama2, LG, 4-gram, synonym & 22.297     & .661   & .995            & .995 & .895          &  .895    & 27.125   & .585 & .986           & .986 & .880 &.880 \\
Llama2, LG, 4-gram, parallel & 22.649     & .637   & .145            & .870 & .211          &  .811    & 25.962   & .683 & .122            & .751 & .165 & .672 \\
Llama2, LG, 4-gram, \textbf{mixing} & 22.430     & .670   & .499            & \textbf{.713} & .440          &  \textbf{.618}    & 26.997   & .648 & .360            & \textbf{.565} & .262 & \textbf{.410} \\  

\bottomrule

\end{tabular}
}

\caption{Main results when LatticeGen (LG) is applied to both the generation and the prompt. All metrics are the lower the better except PMI.  While the generation quality and alignment are degraded, LatticeGen with the proposed mixing scheme successfully protects the true generation from RBS attack to a remarkable degree (measured by max-true-ratio/BERTScore).}
\label{tab:main_gen}
\figvsbottom
\end{table*}

%% file: figure_main.tex
\begin{figure*}
    \framebox[\linewidth]{
    \begin{minipage}{0.98\linewidth}
    \footnotesize
    
    \textbf{Prompt:}
    \textcolor{blue}{Prompt: Aliens have arrived, and ask for a single human to plead humanity's case and save them from extinction. The human is selected through a lottery of the entire human race, and on the day of the drawing, your name is picked... Story:}
    
    \begin{spacing}{1.3}
    \end{spacing}
    
    \textbf{Generated Text ($P_M$):} I could feel my heart rate increase . A cold sweat ran down my back . I could not believe what was happening . My name had just been drawn . Everyone 's names were in a big bowl , with the most common names at the top , to the least common at the bottom
    
    \begin{spacing}{1.3}
    \end{spacing}
    
    \textbf{Generated Text (LG):} 
    \textcolor{blue}{I can see them . They are here to save us from our own destruction , but to watch over us . ''    `` Why have you come ?  What is so important about humans ? '''  `` Humanity has been here since the beginning . They took us by surprise a few years ago .}
    
    \begin{spacing}{1.3}
    \end{spacing}
    
    \textbf{First Round RBS:}  
    Prompt: \RI{Aliens have arrived} \ul{on} the cover of every \RI{single human} \ul{,} and they all have a different colour. Story: \RI{from extinction . The human is selected through a lottery of the entire human race, and on the day of the drawing} is the room with the blue Story: `` We have come in peace . They are not hostile . ''   `` I do n't know ''    `` Why have you come ? \ul{'} \RI{What is so important about humans ? '''  `` Humanity has been here since the beginning . They took us by surprise a few years ago .}

    \begin{spacing}{1.3}
    \end{spacing}
    \textbf{Second Round RBS:}
    Prompt: You\ul{en}stein '\RI{, and ask for a} \ul{meeting} room \RI{to plead humanity's case and save them} `` \ul{the}\ul{int}\ul{.} . ''   The \ul{.}gov \ul{drawing} room . \ul{all} the walls are painted with \ul{you} and you can \ul{your} choice\RI{, your name is picked...} Story: \RI{I can see them . They are here to save us from our own destruction , but to watch over us .}\ul{We} 're here to protect \ul{been} so peaceful and gentle ? ''     `` They 're a threat to us . ''    `` But we were n' \ul{million} species from the

    \end{minipage}
    }
    \captionof{figure}{An example of text generation with LatticeGen, using the configuration of 4-gram, $N$=2 and the mixing scheme. The true tokens are italicized in both rounds of RBS, and the underline indicates that the noise token is mixed from the previous true token. Note that the prompt is also noised by LG.}
    \label{fig:mainexample}
    \figvsbottom
\end{figure*}

%% file: sec_discuss.tex
\section{Limitations}
\label{sec:discuss}

LatticeGen sacrifices generation quality and speed for obfuscation of generated contents. While we show the quality degradation can be alleviated to some degree by using larger $G$-gram unit, it would also cause the inference computation to grow exponentially. An interesting future direction is that, instead of running an inference for all $N^G$ grams, we only select a small portion strategically. 

On the other hand, in this work we focus on protecting the user and the (repeated) beam-search attack from server. There could be other forms of interesting or stronger attacks on the server side (e.g., manual inspection from a human). On the other hand, sharing generation control with client could also endanger the server (e.g., jailbreaking) \citep{liu2023jailbreaking,li2023multistep}. 

Finally, in the current implementation, we lattice-finetune a seperate OPT model for every different lattice configuration, which is space unfriendly. As future work, it would be interesting to explore a unified format of linearized lattice by which a single LLM can process different lattice configurations. 

\section{Broader Impact}
\label{sec:broader}

As stated in \Sref{sec:intro}, in the current user--server interaction paradigm, both the prompt and the generation are raw texts which are completely transparent and accessible to the server. This leaves zero options for users who want to keep the generated text to themselves. On the other hand, the privacy protection offered by today’s LLM providers’ data usage and retention policies is far from enough (detailed in \Sref{appsec:industry}). We propose LatticeGen as a novel protocol for privacy-aware generation with a controlled level of obfuscation. We hope our work could raise awareness for the privacy considerations of generated contents.

%% file: sec_related.tex
\secvsabove
\section{Related Work}
\label{sec:related}
\secvsbelow

Existing work in privacy-aware natural language processing (NLP) \citep{chen21bertprivacy, mcmahan17dprnn} mostly focuses on protecting user data for training \citep{huang-etal-2020-texthide,yue-etal-2023-synthetic} or inference, and the majority of works focus on natural language understanding (NLU) tasks \citep{feyisetan20privacy,xu-etal-2021-utilitarian}. To the best of our knowledge, our work is the first to consider decoding-time privacy for LLM prompted generation on cloud.

\paravs
\paragraph{Lattice in NLP} Lattice \citep{Young2006htkbook} is a graphical structure widely used in structured prediction problems to represent a range of hypotheses. In this work we adopt a simple linear-graph form of lattice which is known as the confusion network \citep{Mangu1999FindingCA}. The lattice structure has found interesting applications in neural NLP models. As a pioneering work, \citet{su2017latticernn} proposes lattice-based RNN encoders for machine translation, where the lattice is generated by merging results from different segmenters. \citet{buckman-neubig-2018-neural} proposes a neural lattice language model, which constructs a lattice of possible paths (segmentations) through a sentence in order to model multiple granularities. Lattice-BERT \citep{lai-etal-2021-lattice} trains LLM to predict a masked portion of a lattice representing possible segmentations of a sentence.
To the best of our knowledge, our work is the first to utilize the lattice structure for privacy-aware generation.

\paravs
\paragraph{Prompt Anonymization} Contemporary and independent of our work, \citet{chen2023hide} proposes to anonymize the named entities (e.g., change \texttt{USA} to \texttt{<GPE>}) in the prompt, and de-anonymize after receiving the generated text from server. In comparison, LatticeGen offers a more general option in that all types of tokens, especially the generated tokens, can be noised.

Due to lack of space, we discuss related work on \textbf{differential privacy}, \textbf{homomorphic encryption} in \Sref{appsec:related}.

%% file: appendix.tex
\clearpage

\appendix

\section*{Supplemental Materials}

\section{Model Training and Inference with Lattice (Server)}
\label{sec:modeltrain}

\paragraph{LLM Finetuning and Inference with the LLG (Linearized Lattice plus G-gram) Format} We now describe how $P_L$ is obtained by finetuning a standard autoregressive LM $P_M$ parameterized by $\theta$ to make next-token predictions with the LLG format(\Sref{sec:preliminary}). We assume access to a public corpus $D$ for finetuning. For simplicity, we focus on the training objective for one length-$T$ sentence $w^d \in D$ and we also assume $N=2$ and $G=1$ (the process for $N>2$ or $G>1$ is highly similar). 

For each data sample $w^d$, we randomly pick another data sample $w^{d'}$ to serve as a 
``parallel'' noise sample, which is used for constructing the noised lattice $W^2_T$ for $w^d$. For time-step $t$, the token in the data sample $w^d$ will be used as the true token $\wtt_t:=w^d_t$, and the token from the parallel sample is used as  the noise token $w^\text{noise(1)}_t:=w^{d'}_t$. To be consistent with the actual generation protocols for LatticeGen, the tokens on each time-step are shuffled.  


The noise generation scheme used by server in the finetuning stage might be different from the scheme used by client in the actual generation. For example, if we use a simple synonym scheme, the perplexity of the synonym scheme during generation will be better. In our implementation we adopt the parallel scheme described above during training because it works well with the proposed mixing scheme (\Sref{sec:def_branch}).


After constructing the noised lattice $W^2_T$, we randomly select $P$ tokens in $w^d$ (we use $P=8$ in our training), and use them as the target next-tokens to finetune the LLM with the LLG format. Denoting their indices as $\{t^1,...,t^{P}\}$, we formulate the following objective:
\begin{equation} \small
\label{eq:latticefinetune}
\begin{split}
        \mathcal{L}_\text{lattice-FT}(w^d, W^2_T;\theta) &= \\ 
&\frac{1}{P} \sum^P_{p=1} \log P_\theta(\wtt_{t^p}|W^2_{t^p-1}[\wtt_{t^p-1}]).
\end{split}
\end{equation}



We now discuss how the server can do efficient LLM inference at time-step $t$. Since $\text{linearize}(W^N_{t-2})$ from the previous time-step $t-2$ is a prefix of $\text{linearize}(W^N_{t-1})$, the server can reuse the saved LLM hidden states\footnote{The \texttt{past\_key\_values} in HuggingFace transformers library.} from the last time-step for the inference of $\{P_L(\cdot|W^N_{t-1}[w^i_{t-1}])\}^N_{i=1}$. However, the server still need to enumerate and inference $N^G$ combinations of the $G$-grams in parallel, and that is the major reason for the slowdown.

\paragraph{Implementation Details} Our model implementation, training and inference utilize the HuggingFace transformers library \citep{wolf-etal-2020-transformers}. We finetune $P_L$ with learning rate of $5\times10^{-5}$ and a batch size of 8 for 3 epochs using the PyTorch \citep{NEURIPS2019_bdbca288} implementation of the AdamW \citep{loshchilov2017decoupled} optimizer. For finetuning of Llama2, we adopt LoRA \citep{hu2021lora}. We perform finetuning of the model under various configurations on one Nvidia A40 GPU.

\section{Auxiliary Framework Description}
\label{appsec:framework}


\begin{figure*}[t]
    \centering
    \includegraphics[width=\linewidth]{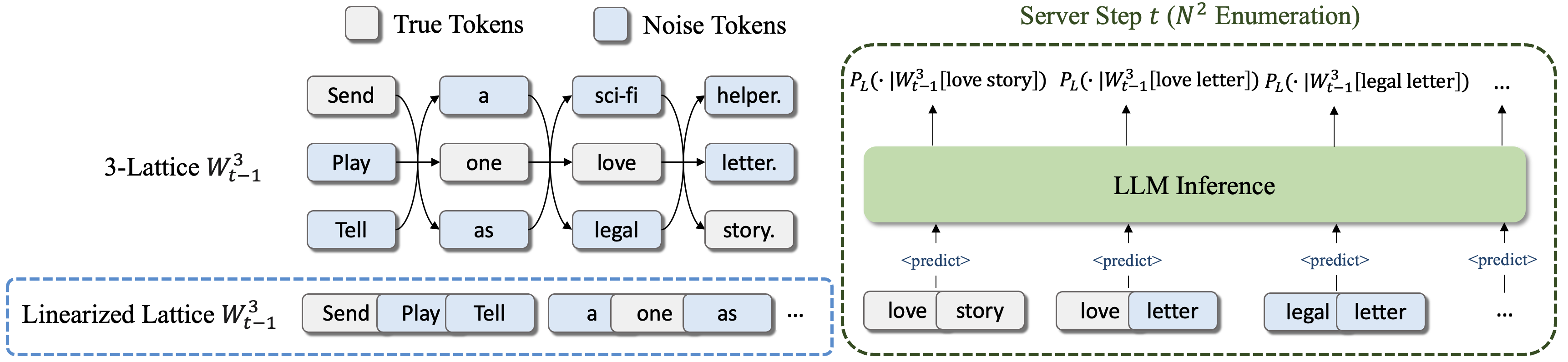}
    \figvsmid
    \caption{An illustration of the server step for $N=3$ and $G=2$. The information of which tokens are the true tokens is only known to the client.}
    \label{fig:serverstep_bigram}
    \figvsbottom
\end{figure*}

\begin{figure}[t]
    \centering
    \includegraphics[width=\linewidth]{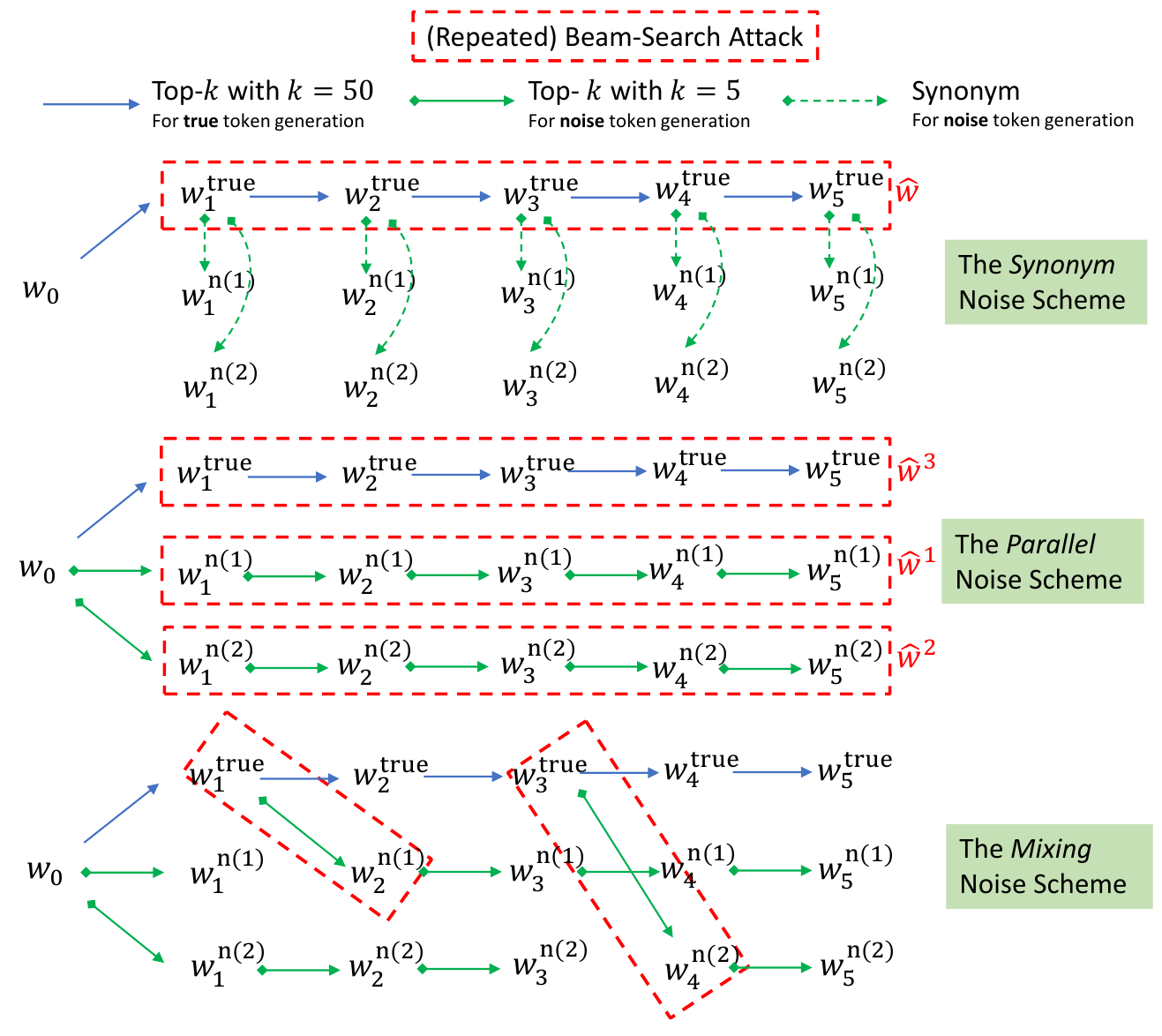}
    \caption{Illustration of different noise schemes under (repeated) beam-search attack. For convenience, the lattice is not shuffled.}
    \label{fig:rbsatk_n3}
\end{figure}

An illustration of the server step for $N=3$ and $G=2$ is provided in Figure \ref{fig:serverstep_bigram}.

An illustration of various noise schemes with a width-3 lattice is provided in Figure \ref{fig:rbsatk_n3}. 

\subsection{Incorporating the Prompt (Client)} 
\label{appsec:prompt}

The prompt $p$ can be easily incorporated by the following. At all time-steps $t$ with $t \leq \text{len}(p)$, instead of sampling $\wtt_{t}$ from $P_L(\cdot|W^N_{t-1}[\wtt_{(t-G)...(t-1)}])$, the client directly sets $\wtt_{t}:=p_{t}$. All other steps in the protocols including the noise token generation continue as normal. In this way, the prompt is also embedded and noised in the lattice.

\subsection{Communication Cost} 
\label{appsec:communicate}

At each time-step, the server needs to send client $N^G$ length-$|V|$ vectors, which could be slow if $|V|$ is large. This can be largely alleviated if the client and server can agree upon a sampling algorithm beforehand. For example, if top-$k$ sampling with $k=50$ is used, then only the logits and indices of the top-50 tokens are needed.

\subsection{The Non-Lattice Baseline}
\label{appsec:nonlattice_baseline}

The training for the non-lattice baseline is a bit similar to the lattice finetuning process described in \Sref{sec:modeltrain}, with the difference that the true tokens are not included in the input. Following the notations in \Sref{sec:modeltrain} with $w^d$ as the data sample, the training objective is formulated as:
\begin{equation}\small
    \mathcal{L}_\text{non-lattice,syn.}(w^d;\theta) = \frac{1}{T} \sum^T_{t=1} \log P_\theta(w^d_t|w^\text{noise}_{0..t-1}),
\end{equation}
where $w^\text{noise}_t$ is randomly set to a synonym of $w^d_t$.
Basically, the model is trained to predict the next true token with a ratio of input tokens noised.

\section{Related Work}
\label{appsec:related}

This section continues from \Sref{sec:related}.

\paravs
\paragraph{Differential Privacy (DP) for LM Training and Inference} 
There are numerous existing works on how to train LLMs with differential privacy~\citep{li2021large,yu2021differentially}, which mostly rely on DP-SGD \citep{dpsgd2016} and limits leakage of private data during training. More related to LatticeGen is a line of work with local DP \citep{xu2020differentially,meehan2022sentence}, which applies discrete noise onto text and can be used to synthesize private text data \citep{yue-etal-2023-synthetic,mireshghallah-etal-2023-privacy}. 

It is not directly clear how these techniques can be adapted for our setting of privacy-aware autoregressive text generation. In comparison, LatticeGen provides a totally different and cooperative approach with the lattice structure and novel defense and attack schemes. 

\paravs
\paragraph{Homomorphic Encryption} There is also a line of work \citep{chen-etal-2022-x} applying techniques from homomorphic encryption~\citep{gentry2009fully} to transformer LM. While they enjoy nice cryptographic guarantees, the induced computational cost is usually huge.

\section{Auxiliary Results}
\label{appsec:auxres}

\begin{table}[t]
\footnotesize
\centering
\addtolength{\tabcolsep}{-3.0pt}
\begin{tabular}{c|ccc}
\toprule
Speed (second/token) & N=1 & N=2 & N=3 \\ 
\midrule
$P_M$ & .013 & / & / \\
LG, Unigram   & /   & .024 (1.84x)   &  .028 (2.15x)  \\ 
LG, Bigram    & /   & .028 (2.15x)   &  .047 (3.62x)   \\ 
LG, 4-gram    & /   & .062 (4.76x)  &  .332 (25.53x)  \\ 
\bottomrule
\end{tabular}
\caption{Generation speed comparison between different systems. For LG, the mixing noise scheme and the OPT model is used. Our implementation is run on a single A40 GPU.}
\label{tab:speed_table}
\end{table}



Similar to Figure \ref{fig:mainexample}, Figure \ref{fig:n2-another-example-with-prompt} shows an example using a different prompt using bigram $N$ = 2. 

On the single A40 GPU we use, LG with bigram units ($N=2$) has a 2x slowdown comparing to $P_M$ (Table \ref{tab:speed_table}, \Sref{appsec:auxres}). Since inference with transformer model benefits from parallel computing, the slowdown should be less significant on servers with stronger computing power.






\section{The Current Privacy Protection Practices in Industry}
\label{appsec:industry}

The privacy protection offered by today’s LLM providers’ data usage and retention policies is far from enough. \footnote{\url{https://opaque.co/announcing-opaqueprompts-hide-your-sensitive-data-from-llms/}} For example, OpenAI’s consumer-facing ChatGPT used to train its models with user input, and also shares user input with third-party providers, and Google’s Bard retains user activity for at least 3 months. As a striking example, employees in Samsung reportedly shared sensitive code with OpenAI during their interaction with ChatGPT. \footnote{\url{https://gizmodo.com/chatgpt-ai-samsung-employees-leak-data-1850307376}} More recently, some of the users' conversations with Bard are mistakenly indexed and accessed by Google search. \footnote{\url{https://venturebeat.com/ai/oops-google-search-caught-publicly-indexing-users-conversations-with-bard-ai/}}

While providers have recently improved their security posture (e.g., OpenAI no longer uses data submitted via its API to train its model), users still can not assume that all sent/received data will be immediately and completely deleted. Rather than regulations, our proposed LatticeGen takes an algorithmic and cooperative approach to give the user advantage and control in privacy protection.

\input{app_examples}

%% file: app_examples.tex
\begin{figure*}
    
    \framebox[\linewidth]{
    \begin{minipage}{0.98\linewidth}
    \footnotesize
    \textbf{Prompt:}
    \textcolor{blue}{Every planet in our solar system has a `` champion '' being that takes on the attributes of the planet itself. The `` champion '' from the sun has created an army to destroy the planets and the 8 ( or 9 ) champions must save the solar system... Story:}
   \begin{spacing}{1.3}
    \end{spacing}
    \textbf{Generated Text ($P_M$):} The planet Mars was known for its reddish color . Mars has a very thin atmosphere , and only a select few had been able to breathe it . But not this man .   This man could breathe anything .   His name is Sol , also known as the Sun .

    \begin{spacing}{1.3}
    \end{spacing}
    \begin{spacing}{1.3}
    \end{spacing}
    \textbf{Generated Text (LG):} 
    \textcolor{blue}{`` There 's nothing you can do , '' I said , running through my head as I saw the soldiers fall . The soldiers were outnumbered , and his army too vast for us to even put up a fight and still lose ? It will be too late ! The champion is here ! ''}
    \begin{spacing}{1.3}
    \end{spacing}
    \textbf{First Round RBS:}  
    Prompt: \RI{Every planet in} \ul{the} \ul{galaxy} \RI{has a `` champion ''} \ul{,} \RI{that takes on the attributes of} \ul{all} of the inhabitantsants \RI{``} \ul{life} \RI{'' from the sun has} \ul{taken} up arms against him .. Story: `` \RI{the} \ul{3} \RI{( or 9 ) champions must save the solar system...} Story: \RI{`` There 's nothing you can do , '' I said , running through my head as I saw the soldiers fall . The soldiers were} too powerful for us ! ''   `` You can try ! ''   `` What ? How ? ''   `` You \ul{not} only have to fight the champion , but his

    \begin{spacing}{1.3}
    \end{spacing}
    \textbf{Second Round RBS:}
    Prompt: A man is \RI{our solar system} 's life is a \RI{being} ul{,} , , , , , \RI{the planet itself. The} . \RI{champion} on Earth \ul{each} \ul{other} to \RI{created an army to destroy the planets and} I ca\RI{8} \ul{other} \ul{I} '\ul{3}m not \ul{are} you \ul{Earth}lingss from Story: The world was in chaos . \ul{say} something ! ''  `` No ! ''   `` \ul{if} we could have stopped him . He was \RI{outnumbered , and his army too vast for us to even put up a fight and still lose ? It will be too late ! The champion is here ! ''}

    \end{minipage}
    }
    \captionof{figure}{Another example of text generation with LatticeGen, using the configuration of 4-gram, $N$=2 and the the mixing scheme. The true tokens are italicized in both rounds of RBS, and the underline indicates that the noise token is mixed from the previous true token. Note that the prompt is also noised by LG.}
     \label{fig:n2-another-example-with-prompt}
    \end{figure*}